\documentclass[journal]{IEEEtran}
\pdfoutput=1
\usepackage{float}
\usepackage{graphicx}
\usepackage[caption=false]{subfig}
\usepackage{textcomp}
\usepackage{xcolor}
\usepackage{float}
\usepackage{multirow}
\usepackage{csquotes}
\usepackage{amsmath,amssymb,amsfonts}
\usepackage{url}
\usepackage{hyperref}
\hypersetup{colorlinks=true,linkcolor=blue,citecolor=blue,filecolor=blue,urlcolor=blue}
\usepackage{tabularx}
\usepackage{booktabs}
\newcolumntype{Y}{>{\raggedleft\let\newline\\\arraybackslash\hspace{0pt}}X}
\newcolumntype{Z}{>{\centering\let\newline\\\arraybackslash\hspace{0pt}}X}

\usepackage{enumitem}
\setlist[itemize]{noitemsep,topsep=0pt}

\newcommand{\tu}{\textsubscript}

\ifCLASSINFOpdf

\else

\fi

\begin{document}

\title{Synthetic ID Card Image Generation for Improving Presentation Attack Detection}

\author{Daniel Benalcazar, Juan E. Tapia ~\IEEEmembership{Member,~IEEE,}, Sebastian Gonzalez,
    and~Christoph Busch,~\IEEEmembership{Senior Member,~IEEE,}\\ 
**The following paper is a pre-print. This manuscript is currently under the Journal review process**     
\thanks{Juan Tapia and Christoph Busch are with the da/sec-Biometrics and Internet Security Research Group, Hochschule Darmstadt, Germany, e-mail: \{juan.tapia-farias, christoph.busch\}@h-da.de.}
\thanks{Daniel Benalcazar and Sebastian Gonzalez, are with the TOC Biometrics Company R\&D Center, Santiago, Chile, email: \{daniel.benalcazar, sebastian.gonzalez\}@tocbiometrics.com}%
\thanks{Manuscript received xxx; revised xx.}}

\markboth{Journal of \LaTeX\ Class Files,~Vol.~14, No.~8, August~2015}%
{Shell \MakeLowercase{\textit{et al.}}: Bare Demo of IEEEtran.cls for IEEE Journals}

\maketitle


\begin{abstract}
    Currently, it is ever more common to access online services for activities which formerly required physical attendance. From banking operations to visa applications, a significant number of processes have been digitised, especially since the advent of the COVID-19 pandemic, requiring remote biometric authentication of the user. On the downside, some subjects intend to interfere with the normal operation of remote systems for personal profit by using fake identity documents, such as passports and ID cards. Deep learning solutions to detect such frauds have been presented in the literature. However, due to privacy concerns and the sensitive nature of personal identity documents, developing a dataset with the necessary number of examples for training deep neural networks is challenging. This work explores three methods for synthetically generating ID card images to increase the amount of data while training fraud-detection networks. These methods include computer vision algorithms and Generative Adversarial Networks. Our results indicate that databases can be supplemented with synthetic images without any loss in performance for the print/scan Presentation Attack Instrument Species (PAIS) and a loss in performance of 1\% for the screen capture PAIS.
\end{abstract}


\IEEEpeerreviewmaketitle


\section{Introduction}
\label{sec:intro}

\IEEEPARstart{T}{he global} pandemic of COVID-19 accelerated the adoption of remote biometric authentication for online services such as e-commerce, digital banking, fintech and document signing. This allowed people to carry on with their normal business activities from home without the risk of spreading the virus. Some services include remotely opening a bank account, something that required physical attendance only a few years back. To access this service, a user only has to validate his/her identity by capturing a selfie and a picture of their ID document. Therefore, remote services made commercial activities more dynamic and accessible for everybody while diminishing the chances of infection.

The downside of remote services is twofold. Firstly, in regions like South America, the accelerated technological leap was too quick for some countries, resulting in difficulties for the national identification systems to catch up with the advancements. For instance, identity documents issued by many countries neither comply with ICAO 9303 standard \footnote{\url{https://www.icao.int/publications/Documents/9303_p4_cons_en.pdf}} nor have NFC chips with the personal information of the individual embedded in their government-issued ID cards. This makes it difficult to automatically verify the genuine nature of an ID card. Secondly, certain individuals might attempt to defraud the remote service by tampering with the photo ID or other fields in the ID card to conceal their identities or impersonate somebody else. In those cases, it is of utmost importance to rely on a system that can check whether an ID card digital image has been manipulated. Even today, there are many templates of ID cards, passports, driver's licenses, and others available to help an attacker produce a high-quality falsified document\footnote{\url{https://gotempl.cc/product-category/id/}}.

Methods that detect fraud in personal ID documents have been presented in recent years  \cite{shi2018docface, shi2019docface+, gonzalez2021hybrid,DBLP:journals/iet-bmt/RajaR022}. Those systems rely on deep learning methods, such as Convolutional Neural Networks (CNN), in order to achieve great detection accuracy~\cite{Zheng2019ASO}. Deep learning systems in general require a large number of examples to train successfully; however, the sensitive nature of ID cards and passports makes it very difficult to acquire the number of images needed. For that reason, in this work, we propose to create synthetic examples of ID cards images to enhance the dataset on which fraud-detection networks are trained. We hypothesise that with the additional synthetic samples, a CNN will produce comparable results to having more real ID cards samples to train with. We explored state-of-the-art Generative Adversarial Networks (GAN), as well as image processing techniques for synthetic image generation.

\begin{figure*}[!htb]
    \begin{center}
        \subfloat[bona fide\label{subfig:bonafide}]{{\includegraphics[width=0.24\linewidth]{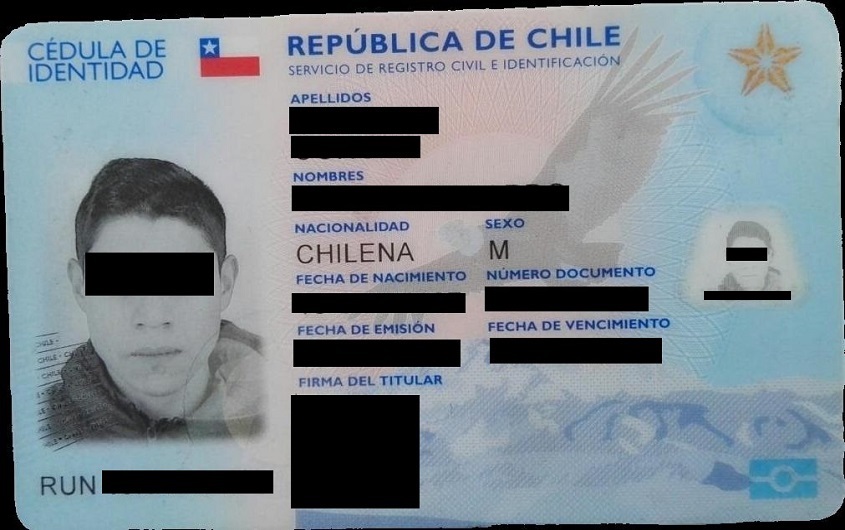} }}%
        \subfloat[composite\label{subfig:composite}]{{\includegraphics[width=0.24\linewidth]{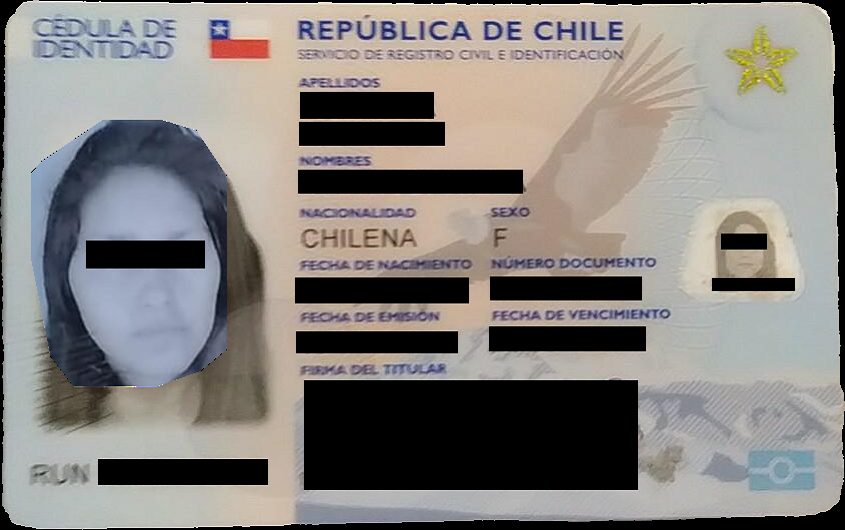} }}
        \subfloat[print\label{subfig:print}]{{\includegraphics[width=0.24\linewidth]{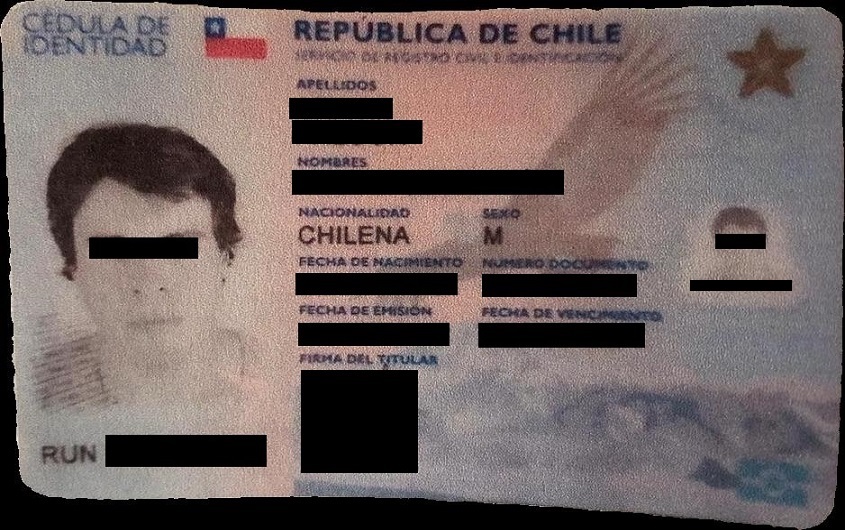} }}
        \subfloat[screen\label{subfig:screen}]{{\includegraphics[width=0.24\linewidth]{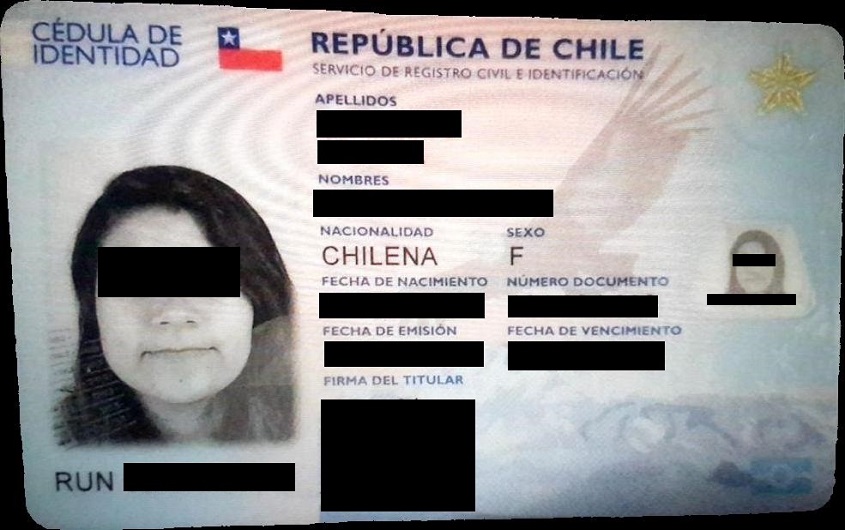} }}
    \end{center}
    \caption{Examples of captured Chilean ID cards of the bona fide class, as well as the presentation attack classes: composite, print and screen.}%
    \label{fig:attacks}%
\end{figure*}

In this work, we consider three kinds of Presentation Attack Instruments (PAI) which are named \enquote{composite (modified manually or automatically)}, \enquote{print} and \enquote{screen}. 
In this way, the four species used in this work for classification will be: bona fide, composite, print, and screen. Examples of the four classes are illustrated in Figure~\ref{fig:attacks}.

To properly increase the number of samples of captured ID card images in our datasets for fraud detection using synthetic images, examples of the four classes must be generated: bona fide, composite, print and screen. We explore GAN models to create more samples of the bona fide class, which can draw more examples of the same distribution of the input images~\cite{viazovetskyi2020stylegan2}, as well as synthesising ID card images from templates with image processing techniques. On the other hand, to create more examples of attack presentation ID card images, we propose adding texture noise from the printing/scanning and screen capture processes. For this purpose, we used GANs, as well as an image processing-based noise isolation and addition technique. Later, for the composite scenario, we also used GANs and automatic splicing techniques that combine two ID card images. Finally, To evaluate the impact of the synthetic images, we trained two MobileNetV2~\cite{sandler2018mobilenetv2} networks (based on Gonzalez et al.~\cite{gonzalez2021hybrid}) using images of Chilean ID cards. We compare training performance with captured images only against training with a combination of captured and synthetic images. It is essential to point out that this work is one of the first to directly synthetic ID card images.

The contributions of this work are as follows:

\begin{itemize}
    \item \textit{ID card generation from templates}: A traditional image processing algorithm is developed, capable of generating fake ID card images from a clear template that resembles the original bona fide and presentation attack scenarios.
    \item \textit{Texture transfer-based presentation attack ID card image generation}: A traditional image processing algorithm capable of generating ID card images faster than GANs-based methods is proposed. This algorithm transfers the noise textures from presentation attack species images to bona fide images. These noise texture templates will be available to other researchers.
    \item \textit{Synthetic ID card generation}: GAN models capable of generating synthetic ID cards that resemble the original bona fide and presentation attack images are developed. Generated images will be made available to other researchers.
    \item \textit{Benchmark evaluation}: The influence of having synthetic ID card images as part of the training dataset of a state-of-the-art ID card image fraud detection network~\cite{gonzalez2021hybrid} is evaluated.
    \item \textit{Analysis}: A comprehensive analysis of the difficulties of synthetic ID card generation and the proposed methods' benefits are provided.
\end{itemize}

The rest of the article is organised as follows: Section~\ref{sec:relate} summarises the related works on generative adversarial networks and image tampering. New methods for generating synthetic images are described in Section~\ref{sec:method}. The experimental framework and results of this work are then presented in Section~\ref{experiments}. We conclude the article in Section~\ref{sec:conclusions}.

\section{Related Work}
\label{sec:relate}

The GAN algorithm was first introduced by Ian Goodfellow et al.~\cite{IAN}. It approaches the problem of unsupervised learning by simultaneously training two deep networks, called Generator $G$ and Discriminator $D$ respectively. These networks compete and cooperate with each other. While the generator creates new instances of the data, the discriminator evaluates them for authenticity. In the course of training, both networks learn to perform their tasks. To learn the generator distribution $p_{g}$ over data $x$, the generator builds a mapping function from a prior noise distribution $p_z(z)$ to data space as $G(z;\theta_g)$. The discriminator $D(x;\theta_d)$, on the other hand,  outputs a single scalar representing the probability that $x$ came from training data rather than $p_{g}$. During training, the parameters of $G$ to minimise $log(1-D(G(z)))$ and the parameters of $D$ to minimise $log(D(x))$ are simultaneously adjusted as if they were following a two-player min-max game with value function $V(G, D)$, specified in Equation~\ref{eqn:gan}.

\begin{equation}
    \label{eqn:gan}
    \begin{split}
        \min\limits_{G} \max\limits_{D} V(D,G) = \mathop{\mathbb{E}}_{x \sim p_{data}(x)}[log(D(x))] \\ + \mathop{\mathbb{E}}_{z\sim p_{z}(z)}[log(1-D(G(z)))]
    \end{split}
\end{equation}

\begin{figure*}[!htb]
    \centering
    \includegraphics[width=\linewidth]{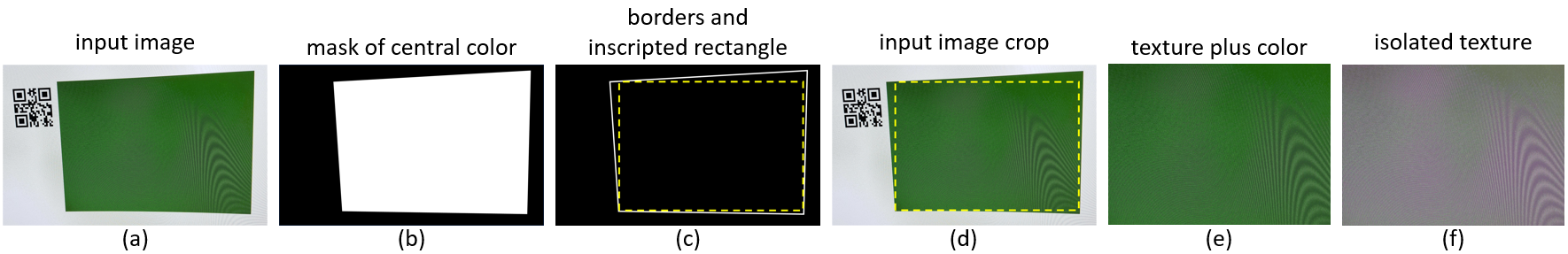}
    \caption{Process of device-noise texture isolation for artificial generation of print and screen attack species.}
    \label{fig:texture_isolation}
\end{figure*}

Karras et al. developed StyleGAN2~\cite{viazovetskyi2020stylegan2, karras2020training} as an extension of the progressive, growing GANs. This approach enables training face generator models capable of synthesising large high-quality images via the incremental expansion of discriminator and generator models from small to large images during the training process. In addition to the gradual growth of the models during training, the StyleGAN2 changes the architecture of the generator significantly. The StyleGAN2 generator no longer takes a point from the latent space as input; instead, two new sources of randomness are used to generate a synthetic image: a standalone mapping network and noise layers. StyleGAN2 introduces the mapping network $f$ to transform $z$ into this intermediate latent space $w$ using eight fully connected layers. This intermediate latent space $w$ can be viewed as the new $z$, $(z^\prime)$. Through this network, a 512-D latent space $z$ is transformed into a 512-D intermediate latent space $w$ applied to face images.

CycleGAN~\cite{chu2017cyclegan} is a technique for training unsupervised image translation models via the GAN architecture using unpaired collections of images from two different domains. Its utility has been demonstrated in a range of applications, including translating seasons in photographs, object transfiguration, style transfer, colorization, and generating photos from paintings. The discriminator models classify $70\times70$ overlapping patches of input images as belonging to the domain or having been generated; the discriminator output is then taken as the average prediction for each patch.

There are some works in the literature that have tackled the problem of fake ID cards detection. For instance, Shi and Jain~\cite{shi2018docface, shi2019docface+} proposed DocFace and DocFace+ to determine the authenticity of a personal ID document by comparison of a face image (selfie) with the photo ID in the ID document. In DocFace+, a kiosk scans the ID document photos or reads the photo from the embedded chip using an NFC reader.

Zheng et al.~\cite{Zheng2019ASO} present a survey and provide an overview on typical image tampering methods, released image tampering datasets, and recent tampering detection approaches. It presents a distinct perspective to rethink various assumptions about tampering clues, which can be discovered by different detection approaches. This further encourages the research community to develop general tampering localisation methods in the future instead of adhering to single-type tampering detection. Most of the analysis was realised using handcrafted images in different domains. Deep learning methods have not been explored in detail.

Albiero et al.~\cite{albiero} present a method for compare selfie images to photo ID images from Chilean ID cards, across adolescence, employing fine-tuning techniques using a private dataset.


Stokkenes et al.~\cite{stokkenes2018biometric}, proposed an online banking authentication system based on features extracted from faces using bloom filters. This information is encoded and used as a key for accessing banking services.

Perera et al.~\cite{perera2019face} proposed an active authentication system that attempts to continuously monitor user identity after access has been initially granted. A similar approach has recently been reported by Fathy et al.~\cite{fathy2015face}.

Arlazarov et al.~\cite{MIDV-500} presented a tiny ID cards dataset containing 500 video clips of 50 different identity document types. This dataset was one of the first made publicly available for identity document analysis and recognition in video stream. Additionally, the paper presents three experimental baselines obtained using the dataset: face detection accuracy, separate text fields OCR precision for four major identity document field types, and identity document data extraction from video clips.

On a different approach, Gonzalez et al.~\cite{gonzalez2021hybrid} proposed a hybrid two-stage classification system that checks if the entire Chilean ID card was tampered or modified. In that work, the authors analysed the primary sources of fraud as image composition and image source tampering. This paper is relevant because it presents results on genuine transactions of a remote verification system with 24,778 images distributed on bona fide, composite, printed and display ID card images.

Zhu et al.~\cite{zhu} propose a method that indicates whether the photo ID or fields in the ID card image have been altered or replaced by digital or handcrafted means. On the other hand, if the source of the ID card comes from a printing/scanning process or the image was captured from a digital screen, it means that it does not come from the original plastic document, and alterations could have been made beforehand.

Mudgalgundurao et al.~\cite{DBLP:journals/iet-bmt/RajaR022} proposed a method to detect fake German ID cards and residence permits using pixel-wise supervision based on DenseNet. This technique enables the method to leverage minute cues on various artefacts, such as moiré patterns and artefacts left by the printers. The authors present the baseline benchmark using different handcrafted and deep learning models on a newly constructed in-house database obtained from an operational system consisting of 886 users with 433 bona fide, 67 print and 366 display attacks.

Regarding the training and testing data, due to the privacy concerns of the ID cards, the databases for research and commercial purposes are difficult to obtain. So far, there is a limited number of databases available: the Chilean Sequestered ID card database used in~\cite{gonzalez2021hybrid}, Public-IvS where the ID card images are cropped from the existing images from CASIA-IvS~\cite{zhu}, and the German ID card and resident permit images~\cite{DBLP:journals/iet-bmt/RajaR022}. Furthermore, the MIDV-500 dataset~\cite{MIDV-500}, which contains video clips of 50 different identity document types, including 17 types of ID cards, 14 types of passports, 13 types of driving licenses is also available.
\vspace{-0.3cm}

\section{Methods}
\label{sec:method}

This section describes the implementation of two MobileNetV2 networks used as an ID card fraud-detection system, and the generation of synthetics images using three different methods: templates, texture transfer and GAN models for ID cards.

The goal of the synthetic images is to be as close as possible to the captured images of the Chilean ID card database; however, these techniques can be easily extended to other countries. The Chilean ID card dataset used in this work is explained in Section~\ref{sec:database}. Each image on the dataset was segmented using the network proposed by Lara et. al.~\cite{lara2021towards} in order to remove the background pixels and force the fraud-detection network to learn features only from the pixels inside of the  ID card. Figure~\ref{fig:attacks} show examples of the segmented ID card images.

\subsection{Fraud-Detection Network}

To detect tampering in ID cards and evaluate the impact of synthetic images, we implemented two MobileNetV2~\cite{sandler2018mobilenetv2} approaches based on~\cite{gonzalez2021hybrid}. The first network detects composition tampering, and was trained to perform a binary classification task between bona fide and composite images. The network input size is $224\times224$ pixels and the number of filters in the network was increased using an alpha parameter of 1.4. For the second network, the system must identify the source of the image among bona fide presentations, and print/screen attack presentations. In this case, the network input is $448\times448$ pixels, and the alpha parameter was also raised to 1.4. The greater the  resolution better represents the fine-grain device noise, which would be lost in a lower resolution.
\vspace{-0.3cm}

\subsection{Image generation from templates}

\begin{figure}[!htb]
    \centering
    \includegraphics[width=0.7\linewidth]{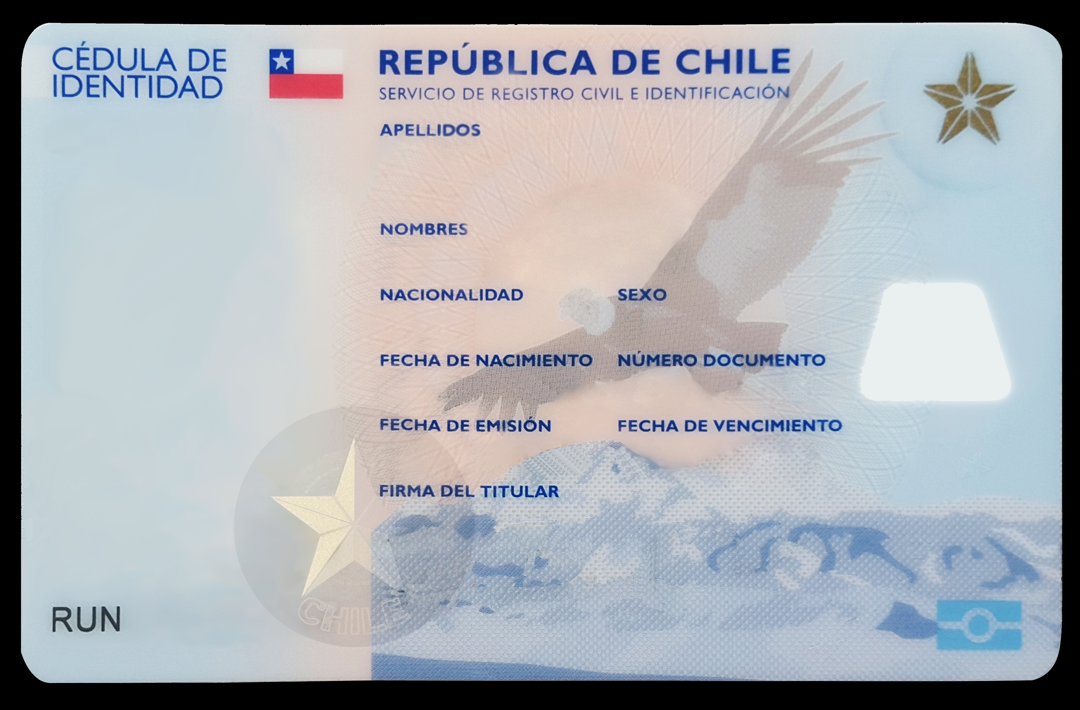}
    \caption{Example of the Chilean ID card template.}
    \label{fig:template}
\end{figure}

In this paper, deterministic image-processing techniques that can automatically generate ID cards from templates were also analysed. To create the templates, we started by selecting one high-quality bona fide example of a Chilean ID card. Then we straightened out the image and emptied all the fields, signatures and photos using Photoshop's cloning tool. The result is a clean background template ready to be automatically filled, as shown in Figure~\ref{fig:template}. 
After that, a series of algorithms were developed to select random faces and signatures, as well as random names, dates and alphanumeric characters, to fill the templates with random information. The face images are selected from the bona fide probe FERET dataset\footnote{\url{https://www.nist.gov/itl/products-and-services/color-feret-database}}, while the signatures originated from the Kaggle dataset of handwritten signatures\footnote{\url{https://www.kaggle.com/datasets/divyanshrai/handwritten-signatures}}. During the information generation process, we took special care to match the gender of the photo with that of the first name and the gender field in the ID card. We also generated plausible dates and numbers for each field. We draw from a dictionary containing Latin American countries' most familiar names to create random names and surnames.  

Afterwards, we applied a few post-processing steps. We first used a small random variation in the Hue, Saturation, and Value (HSV) channels to slightly change the colour of the ID cards. This method is intended to mimic changes in the light source. Then, we applied a random perspective change to resemble the differences in perspective found in bona fide ID card images. This is due to the difficulty of holding the camera perfectly parallel to the document when the user takes pictures of their ID cards
. This is replicated using a random protective transformation. Finally, we confirm that the background of the image is entirely black (R=0, G=0, B=0), resembling the result from the segmentation network~\cite{lara2021towards}.

Since the face image dataset contains 2,069 images, we decided to create synthetic ID cards in batches of this number. In this way, every batch will have the same faces in the same order but with different signatures, text fields, colours and perspectives. In the remainder of this paper, we will refer to this method as \enquote{Templates}, for short.
\vspace{-0.3cm}

\subsection{Image generation from transferable textures}
\label{subsec:palette}

This proposed image-processing technique isolates an image's texture and transfers it to another. This process was applied to generate the print and screen scenarios. In this way, an automatic method could replace the labour-intensive process of capturing bona fide images, printing or displaying them on a screen, and re-capturing them to create replay attacks. 

\begin{figure}[H]
    \centering
    \includegraphics[width=0.9\linewidth]{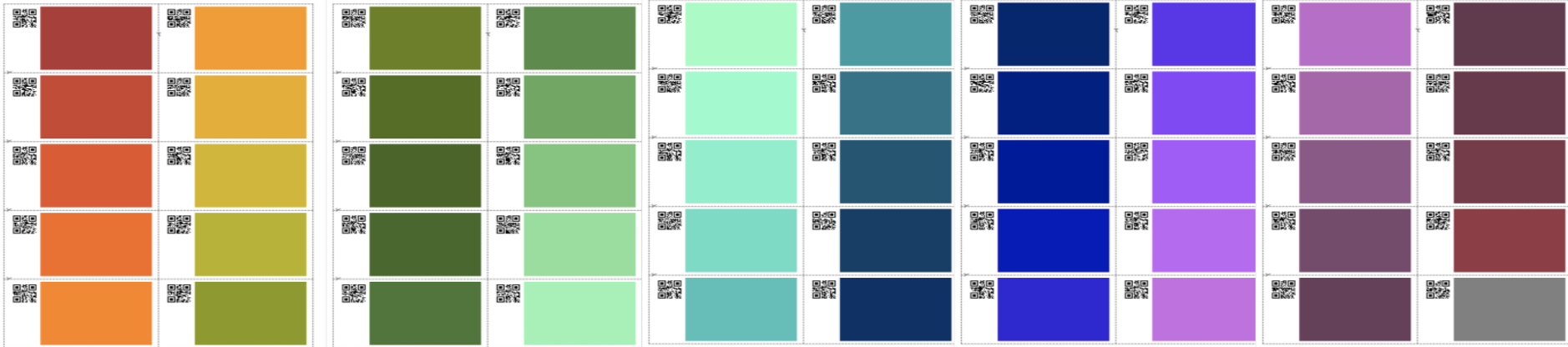}
    \caption{Example of the 50 colours palette used to create artificial textures.}
    \label{fig:palette}
\end{figure}

The primary motivation behind this method is the following: if we have a bona fide image, as well as the presentation attack version of the same image aligned pixel by pixel, in that case, the mathematical subtraction between the two images will cancel out the image information leaving only the differences caused by the PAI. 
However, the print/screen version is unavailable; therefore, we must capture this artefact from other images as a colour palette. After that, it is possible to apply this new texture to the print and screen scenarios. This technique will help us to isolate the noise texture left by the printing/screen or displaying processes. Then, the isolated texture can be added to any other image to tamper with the desired attack noise scenario artificially. To simplify the alignment process, we chose a palette of 50 solid colour images, as shown in Figure~\ref{fig:palette}. Each colour has a QR code on the left side that identifies it.

To isolate the texture on each image of that dataset, we first read the included QR code to identify the original colour. Then, the colour rectangle is segmented, eliminating the QR code and the background. This process is illustrated in Figure~\ref{fig:texture_isolation}. To obtain the segmentation mask (Figure~\ref{fig:texture_isolation}b), we first sampled a $200\times150$ pixel patch around the central pixel of the image and computed the mean and standard deviation in the HSV space. Then, masked all the pixels in the image with colour values around the mean HSV colour of the central patch plus minus four standard deviations. A morphological closing operation is applied to the mask to fill any gaps. Then, the borders of the mask are obtained and the inscribed rectangle is computed, which is used to crop the image. Therefore, at this point, we have an image that only contains pixels with the main colour and the texture of interest (Figure~\ref{fig:texture_isolation}e). Finally, we isolate the texture by subtracting the colour indicated by the QR code and save the texture to create a dataset of 5,000 print and 5,000 screen isolated textures. In Figure~\ref{fig:texture_isolation}f, a constant RGB value of (128, 128, 128) was added to the isolated texture for visualisation purposes only.

In the end, a dataset of 10,000 isolated textures is obtained, where any texture can be added to any bona fide image to artificially generate the tampering effect. This process consists of cropping the isolated texture with the image size of the bona fide image, and then applying a pixel-whise mathematical addition between the texture and the bona fide image, as illustrated in Figure~\ref{fig:texture_addition}. In this work, we applied the transferable texture technique to generate attack presentations from captures ID card images and Templates. Since the textures come from captured images, it is expected that fraud-detection networks would learn to identify the noise patterns as the corresponding attack scenario. In the sections below, we will refer to this method as \enquote{Textures}.

\subsection{Image generation from GANs}

The following GAN networks were used in this work: StyleGAN2-ADA~\cite{viazovetskyi2020stylegan2, karras2020training} and CycleGAN~\cite{chu2017cyclegan}. The later translates images of domain A into domain B, so examples of the two domains are needed during training. In other words, one must guide the training so that the network can translate base images into the desired domain. For CycleGAN, datasets A and B are unpaired. 
On the other hand, StyleGAN2 is capable of generating new examples of the desired domain, without the need for base images~\cite{viazovetskyi2020stylegan2}. To be more precise, StyleGAN2 takes a random vector in the latent space and transforms it into an image of the desired domain, using the distributions learnt during training. Therefore, these two networks are useful for different purposes expanded below. 

StyleGAN2-ADA was used to generate more examples of bona fide ID card images, as well as the three attack presentation species considered in this work: composite, print and screen. Therefore, the class-conditional version of SyleGAN2-ADA was used to generate examples of the four classes in a single training session. The main advantage of SyleGAN2-ADA is that it is optimised to learn from a dataset with limited examples~\cite{karras2020training}. However, synthesising ID card images from scratch is a difficult task to learn since the network has to create faces, signatures, alphanumeric characters, symbols, device noise and perspective changes all at once.

Conversely, CycleGAN was trained to create more examples of the print and screen classes from the bona fide class, using domain adaptation. In this way, the network is meant to learn and mimic the fine-grain textures left by paper, inks and printer devices, as well as pixels and aliasing left by screen displays. Therefore, the difficulty of this is smaller than the previous one since faces, signatures, and characters come from the base images, and the network has to focus only on generating device noise textures.
\vspace{-0.3cm}


\section{Metrics}

This section describes all the metrics that are used to compare the performance of the proposed methods, both for the image generation and Presentation Attack Detection (PAD) task.

\subsection{Frechet Inception Distance}

One of the difficulties with GAN algorithms, and in particularly when applied to ID card images or biometrics in general, is how to meaningfully assess quality of the resulting (synthesised) images.  Only recently, a suite of qualitative and quantitative metrics have been developed to assess the performance of a GAN model based on the quality and diversity of the generated synthetic images~\cite{SalimansGZCRC16, heusel2017fid, XU}. Such proposed of such metrics are: The Inception Score (IS)~\cite{SalimansGZCRC16}, Frechet Inception Distance (FID)~\cite{heusel2017fid} and Perceptual Path Length (PPL)~\cite{karras2020training}. These metrics allow to compare results from different GAN models. The FID score was used in this work to measure the objective quality of the ID cards synthetic images.

Frechet Inception Distance (FID) compares the similarity between two groups of images A and B. First, to compute the FID, all images from set A and set B have to be processed by an Inception v3 network~\cite{pouyanfar2017inception}, pre-trained on ImageNet~\cite{deng2009imagenet}. Then the 2,048 feature vector of the Inception-v3 pool3 layer is stored for each image. Finally, the distributions of A and B in the feature space are compared using Equation~\ref{eq:fid}, where $\mu_A$ and $\mu_B$ are the mean values of the distributions A and B, respectively, and $\Sigma_A$ and $\Sigma_B$ are the covariances of the two distributions.  

\begin{equation}\label{eq:fid}
    FID = \| \mu_A - \mu_B \|^2 + Tr \left( \Sigma_A + \Sigma_B -2  ( \Sigma_A \cdot \Sigma_B ) ^{1/2} \right)
\end{equation}

\subsection{Detection Performance evaluation}

The detection performance of biometric PAD algorithms is standardised by ISO/IEC 30107-3\footnote{\url{https://www.iso.org/standard/67381.html}}. The most relevant metrics for this study are: Attack Presentation Classification Error Rate (APCER), Bonafide Presentation Classification Error Rate (BPCER) and BPCER\tu{AP}. Those metrics determine the the error rates when classifying an instance between bona fide and the different Presentation Attack Instrument Species (PAIS).  

The APCER metric measures the percentage of attack presentations incorrectly classified as bona fide, for each different PAI. To evaluate an entire system, the worst-case scenario is considered. The computation method is detailed in Equation~\ref{eq:apcer}, where the value of $N_{PAIS}$ corresponds to the number of attack presentation images, $RES_{i}$ is $1$ if the $i$th image is classified as an attack, or $0$ if it was classified as a bona fide presentation.

\begin{equation}\label{eq:apcer}
    {APCER_{PAIS}}=1 - \frac{1}{N_{PAIS}}\sum_{i=1}^{N_{PAIS}}RES_{i}
\end{equation}

On the other hand, the BPCER metric measures the proportion of bona fide presentations wrongly classified as attacks. BPCER can be computed using Equation~\ref{eq:bpcer}, where $N_{BF}$ is the amount of bona fide presentation images, and $RES_{i}$ takes the same values described in the APCER metric. The two metrics together determine the performance of the system, and they are subject to a specific operation point. 

\begin{equation}\label{eq:bpcer}
    BPCER=\frac{\sum_{i=1}^{N_{BF}}RES_{i}}{N_{BF}}
\end{equation} 

Finally, to analyse the system performance on an specific operating point, BPCER\tu{AP} and the Equal Error Rate (EER) are used. The later is the operating point where APCER and BPCER are equal. This operating point corresponds to the intersection with the diagonal line in a Detection Error Trade-off (DET) curve, which is also reported for all the experiments. On the other hand, the BPCER\tu{AP} is the BPCER value when the APCER is $100/AP$. In this work, we evaluate BPCER\tu{10}, BPCER\tu{20} and BPCER\tu{100}, which correspond to APCER values of 10\%, 5\% and 1\% respectively.


\section{Databases}
\label{sec:database}

This section describes all the datasets used in this work. The TOC Biometrics company provided the images utilised in~\cite{gonzalez2021hybrid} for research purposes only. A dataset of 9,286 bona fide Chilean ID cards was used. This database was used as a baseline and the starting point to create composite, print and screen images. As mentioned in Section~\ref{sec:method}  a dataset of 10,000 transferable textures was also created. Additionally, we synthesised a significant number of ID card images for the purposes of data augmentation. These datasets are detailed in this section.
\vspace{-0.3cm}

\subsection{Database organisation}

For this work, a significant amount of bona fide images of Chilean ID cards were captured, which complies with ICAO standards. Then, we manually created presentation attack versions of those images for the composite, print and screen species. For the screen species, images of the bona fide ID cards were displayed on monitors, tablets and smartphones, and then images were re-captured using other Android and iPhone smartphones. The resulting images thus have the texture noise of the pixels and aliasing from the displayed devices. Additionally, for the print species, PDF documents containing eight bona fide ID cards images per page were printed in plain and glossy paper. Then, the individual ID cards were cut off with scissors and placed over various surfaces. Afterwards, each ID card was photographed using the same Android and iPhone smartphones used in the screen PAI. The resulting images contain artefacts from the type of paper, inks and the printing/scanning process. Finally, for the composite scenario, the face and fields of the print ID cards were cut and pasted over other print ID cards in order to create fake ID cards. This process was made manually with scissors, and automatically with digital splicing techniques.

Furthermore, as we mentioned before, the background of all captured images was removed using semantic segmentation. This background removal helps fraud-detection networks to focus on the ID card's contents instead of its surroundings.

Table~\ref{tab:chl2} shows the final number of captured images for the bona fide, composite, print and screen classes, as well as test, train and validation partitions. For consistency, the test and validation partitions in this table will be used to evaluate all the fraud-detection networks in this work. We will refer to these partitions as \emph{Chl-test} and \emph{Chl-validation} for short. 

\begin{table}[!htb]
    \scriptsize
    \centering
    \caption{Captured images of Chilean ID cards.}
    \label{tab:chl2}
    \begin{tabular}{lrrrrr}
        \hline
        \textbf{Class} & \textbf{Test} & \textbf{Train} & \textbf{Validation} & \textbf{Total} \\
        \hline
        Bona fide      & 1,842 &  5,613 & 1,831 &   9,286 \\
        Composite      & 1,762 &  6,251 & 2,067 &  10,080 \\
        Print          & 2,322 &  6,893 & 2,299 &  11,514 \\
        Screen         & 1,538 &  4,542 & 1,517 &   7,597 \\
        \hline
        \textbf{Total} & 7,464 & 23,299 & 7,714 & \textbf{38,477} \\
        \hline
    \end{tabular}
\end{table}

In addition to capturing ID card images, we captured a dataset to help isolate the texture noise from the source material, as in the attack species print and screen. The dataset for texture study was captured over a palette of 50 solid colours with an identifying QR code on the left side, as shown in Figure~\ref{fig:palette}. PDF pages containing 10 colours each were printed in several printers, both in plain and glossy paper. Then, several images of each individual colour were captured using smartphones of different brands, at three times of the day (day, afternoon and night) at distances between 15 and 30cm and with portrait and landscape orientations. We captured 2,500 images printed on plain paper and 2,500 images on high-quality glossy paper. The later is meant to resemble the light reflection of the plastic material of authentic ID cards.

\begin{figure*}[!t]
    \centering
    \includegraphics[width=\linewidth]{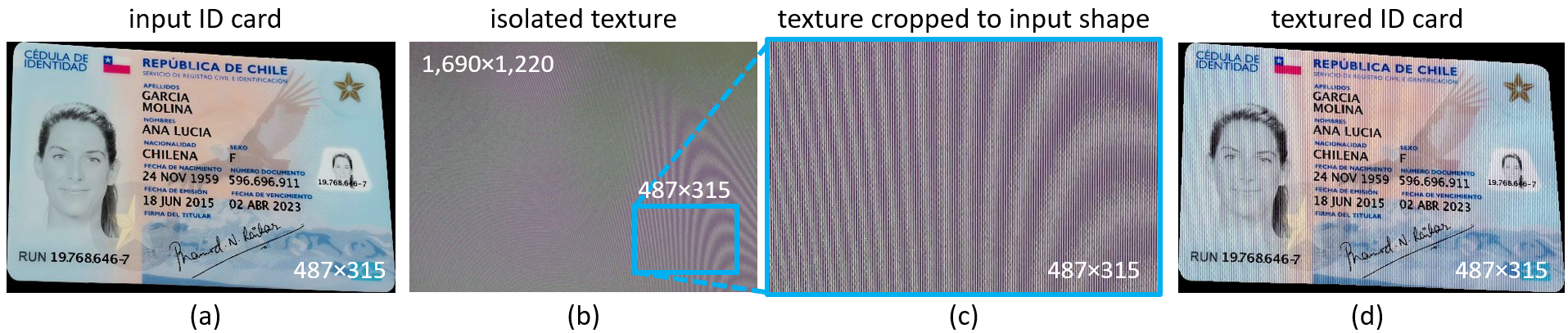}
    \caption{Process of device-noise texture addition. A random isolated texture is selected from the set of 5,000 available textures per PAI. The texture is cropped with the shape of the input image at a random location. The output textured ID card is the pixel-wise addition between the input ID card and the cropped texture. In this figure, an ID card generated by the Templates method is used (no real data), thus it is not blacked out.}
    \label{fig:texture_addition}
\end{figure*}

Similarly, the 50 colours in the palette were displayed on several monitors, tablets and smartphone screens. Images of those projected colours were captured using the same smartphones as above and with the same orientations and times of the day. In total 5,000 images were captured for the screen species. The device noise texture of the 10,000 captured images was isolated with the method described in Section~\ref{subsec:palette} to obtain a dataset of 10,000 transferable textures. Both images and textures will be made freely available \footnote{under acceptance}. The diversity of printers, screens and phones used to capture this dataset makes transferable textures a general-purpose method applicable for other fields. Figure \ref{fig:texture_addition} shows an example of texture application.

Table~\ref{tab:texture} displays the number of images and isolated textures in this dataset. This dataset was used for image processing algorithms. Therefore train, test and validation partitions were not needed. 
\vspace{-0.4cm}

 \begin{table}[!htb]
     \scriptsize
     \centering
     \scriptsize
    \caption{Transferable texture dataset}
     \label{tab:texture}
     \begin{tabular}{lrr}
        \hline
         \textbf{Class} & \textbf{Images} & \textbf{Textures} \\
        \hline
         Bona fide      &              50 &               0 \\
         Print-A        &           2,500 &           2,500 \\
         Print-B        &           2,500 &           2,500 \\
         Screen         &           5,000 &           5,000 \\
         \hline
         \textbf{Total} & \textbf{10,050} & \textbf{10,000} \\
         \hline
     \end{tabular}
 \end{table}
\vspace{-0.3cm}

\subsection{Training Set of Captured and Synthetic Images}

\begin{figure*}[!htb]
    \centering
    \includegraphics[width=1.0\linewidth]{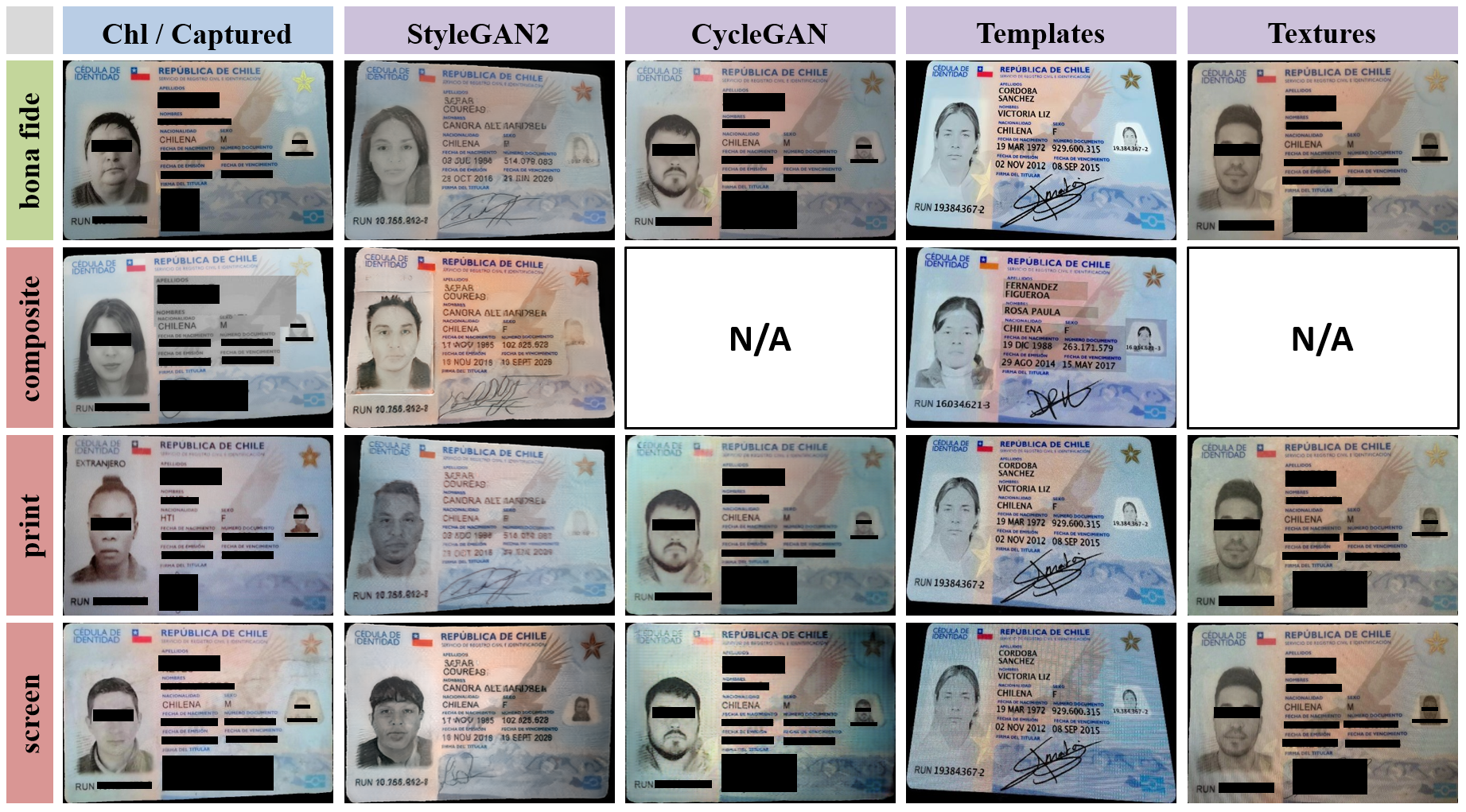}
    \caption{Examples of captured and synthetic images generated with all the proposed methods. The rows and columns are colour coded. Tampered scenarios are in red and untampered in green. Captured images are in blue, whereas synthetic images are in purple. All sensitive information has been blacked out. Fields of StyleGAN2 and Templates are not covered since they contain random data of nonexistent subjects. N/A represent methods not used to generate the particular images.}
    \label{fig:all_methods}
\end{figure*}

Table~\ref{tab:train} shows the quantity of images used for the training of fraud-detection networks. This table includes captured and synthetic images since a combination of both is used to train the networks. Synthetic images were generated using all the methods described in Section~\ref{sec:method}. For a fair evaluation, all trained fraud-detection networks use the same \emph{Chl-test} and \emph{Chl-validation} images described in Table~\ref{tab:chl2}. Examples of captured and synthetic images generated with the proposed methods are presented in Figure~\ref{fig:all_methods}.

\begin{table}[!htb]
    \caption{Train set of captured and synthetic images. SGAN2 represents StyleGAN2.}
    \centering
    \scriptsize
    \setlength{\tabcolsep}{4pt}
    \begin{tabular}{lrrrrrrrr}
        \hline
        \textbf{Class} & \textbf{Chl-A} & \textbf{Chl-B} & \textbf{Chl-C} & \textbf{SGAN2} & \textbf{C-GAN} & \textbf{Templ.} & \textbf{Text.} \\
        \hline
        Bona fide & 2,807 & 2,806 & 2,806 & 3,000 &     0 & 3,104 &     0 \\
        Composite & 3,126 & 3,125 &     0 & 3,000 &     0 & 3,104 &     0 \\
        Print     & 3,447 & 3,446 &     0 & 3,000 & 2,806 & 3,104 & 2,806 \\
        Screen    & 2,271 & 2,271 &     0 & 3,000 & 2,806 & 3,104 & 2,806 \\
        \hline
    \end{tabular}
    \label{tab:train}
\end{table}

The training set of the Chilean ID cards was split into two halves, \emph{Chl-A} and \emph{Chl-B}, in order to perform the proposed experiments in Section~\ref{experiments}. Also, a copy of bona fide \emph{Chl-B} was created and named as \emph{Chl-C}, as the base images for the CycleGAN and texture transfer methods.

For synthetic images, we generated around 3,000 images per class. This represents the average number of images per class in \emph{Chl-A}. 


The Template images were created from a dataset of 3,104 faces. The Templates method can create the bona fide and composite classes by itself; however, for the print and screen classes we translated the template images using the Textures method.

The process for StyleGAN2 was simple, since it can produce any desired number of images. Thus, exactly 3,000 images per class were generated. CycleGAN and Textures translate bona fide images into the attack instrument species print and screen. The base images were the 2,806 bona fide images from \emph{Chl-C}; they were translated to create 2,806 print and 2,806 screen images using CycleGAN and Textures separately.

Examples of all the types of generated images can be seen in Figure~\ref{fig:all_methods}. For captured and StyleGAN2 images, random examples of each class are shown. This is also true for composite images of the Templates method. However, for CycleGAN, Templates, and Textures, the base image, shown on the bona fide row, is translated to generate the print and screen PAIs in the rows below. As mentioned above, neither CycleGAN nor Textures can generate bona fide images, so the bona fide images presented in those two columns are in fact, bona fide images from \emph{Chl-C}.

\section{Experiments and Results}
\label{experiments} 

The experiments described in this section aim to compare the similarity score between captured and synthetic images, as well as evaluating the impact of using synthetic images while training fraud-detection networks. In this work, we assess two MobileNetV2 networks, one that detects composition tampering, and another that detects source alteration.

All evaluations are oriented to test the predictive value of adding synthetic images to the training set under the following scenario: suppose there are $N$ captured images in the train set, which we would like to double. We are interested in studying if a combination of $N$ captured plus $N$ synthetic images would yield similar results to capturing $2 \times N$ images. This is the reason why the train set of captured Chilean ID cards is divided into two halves \emph{Chl-A} and \emph{Chl-B}, as described in Table~\ref{tab:train}. All experiments are explained in detail in this section.

\subsection{Experiment 1 - Generation evaluation}
\label{subsec:exp1}

This experiment assesses the similarity between captured and synthetic images using the FID metric. For this evaluation, we compare each class (bona fide, composite, print and screen) of the test set of captured Chilean ID cards (\emph{Chl-test}) against the corresponding class of synthetic images generated by the proposed methods (StyleGAN2, CycleGAN, Templates and Textures). This evaluation determines how similar captured and synthetic images are for each generation method. Additionally, we compute the FID score between \emph{Chl-test} and \emph{Chl-validation}. This evaluation gives a baseline FID value for captured images under the same conditions. 

Table~\ref{tab:fid} shows the results of the FID scores. The smallest values in the table were obtained when comparing two sets of captured images: \emph{Chl-test} and \emph{Chl-validation}. This comparison has an average FID of 5.45. This represents the best possible achievable performance for a generation method---the further the FID from 5.45, the less resemblance between synthetic and captured images. 

\begin{table}[!htb]
    \caption{FID Scores Computed Between The Proposed Generation Methods and Chl-test.}
    \centering
    \scriptsize
    \setlength{\tabcolsep}{4pt}
    \begin{tabular}{lrrrrr}
        \hline
        \textbf{Class} & \textbf{Chl-val} & \textbf{StyleGAN2} & \textbf{CycleGAN} & \textbf{Templates} & \textbf{Textures} \\
        \hline
        Bona fide        & 4.54 & 20.33 &     - & 36.49 &     - \\
        Composite        & 6.61 & 20.79 &     - & 36.07 &     - \\
        Print            & 4.02 & 21.28 & 13.07 & 30.56 & 13.75 \\
        Screen           & 6.63 & 27.31 & 22.79 & 55.17 & 25.20 \\
        \hline
        \textbf{Average} & 5.45 & 22.43 & 17.93 & 39.57 & 19.48 \\
        \hline
    \end{tabular}
    \label{tab:fid}
\end{table}

StyleGAN2, CycleGAN, and Textures obtained similar FID values around 20. Among them, CycleGAN got the best score despite the success of StyleGAN2 for other image domains, such as human faces~\cite{viazovetskyi2020stylegan2}. We attribute the excellent performance of CycleGAN to the fact that it starts from a base image with a face, a signature and alphanumeric characters already on it. Therefore, this network must only replicate the device noise. For the same reason, Textures got a good score as well. On the other hand, StyleGAN2 has to generate a face, a signature, letters and numbers from scratch, along with perspective changes and device-generated noise.

The worst performance was obtained by the Templates method. We hypothesise that the main reason for this is an excessive range in the random projective transformation. From Figure~\ref{fig:all_methods}, it can be observed that the Templates method had more extreme perspective changes concerning captured and the other synthetic images. Additionally, the random colour shift used could not represent the nature of the distribution of captured ID cards. Further studies are needed to optimise the random projective transformation and colour shift in order to reduce FID.

When comparing the different classes in Table~\ref{tab:fid}, the screen attack was the hardest to replicate by all methods. However, the print scenario produced the smallest FID scores among the synthetic images. CycleGAN achieved an FID of 13.07, while Textures obtained an FID of 13.75. Both values are close to the FID range among captured images. This result means that for the print attack with CycleGAN and Textures, the distribution of synthetic images is very similar to that of captured images.
\vspace{-0.3cm}

\subsection{Experiment 2 - composition detection network}
\label{subsec:exp2}

A binary MobileNetV2 network detects whether an ID card image presents signs of composition or not. We trained and tested this network using only those methods capable of generating composite images, which are StyleGAN2 and Templates. In total, we trained four versions of this network using the following combinations of datasets from Table~\ref{tab:train}.

\begin{enumerate}
    \item Captured: \emph{Chl-A} + \emph{Chl-B} (11,864 images)
    \item StyleGAN2: \emph{Chl-A} + StyleGAN2 (11,933 images)
    \item Templates: \emph{Chl-A} + Templates (12,141 images)
    \item Combined: \emph{Chl-A} + StyleGAN2 + Templates (18,141 images)
\end{enumerate}

Each training was designed to have a similar number of images in the train set, except for the last one, which evaluates the effect of adding images from all the available domains. The trained models were evaluated with the validation and test sets of Table~\ref{tab:chl2} using PAD metrics for each trained model.
\vspace{-0.2cm}

\subsection{Experiment 3 - source detection network}
\label{subsec:exp3}

The second multi-class MobileNetV2 network classifies whether the ID card image came from a bona fide source or if it was tampered with by printing/scanning or screen-capturing PAIS. Therefore, the network classifies into three classes: bona fide, print and screen. In this case, we tested all the methods described in Section~\ref{sec:method}, since they can all mimic source tampering. We trained the network several times with the following combinations of the train sets described in Table~\ref{tab:train}, trying to utilise a similar number of images.

\begin{enumerate}
    \item Captured: \emph{Chl-A} + \emph{Chl-B} (17,048 images)
    \item StyleGAN2: \emph{Chl-A} + StyleGAN2 (17,525 images)
    \item Templates: \emph{Chl-A} + Templates (17,837 images)
    \item CycleGAN: \emph{Chl-A} + \emph{Chl-C} + CycleGAN (16,943 images)
    \item Textures: \emph{Chl-A} + \emph{Chl-C} + Textures (16,943 images)
    \item Combined: \emph{Chl-A} + \emph{Chl-C} + StyleGAN2 + CycleGAN + Templates + Textures (43,673 images)
\end{enumerate}

All trained models use the validation and test sets of Table~\ref{tab:chl2} for consistency. Finally, the PAD scores are also computed for the evaluation and comparison of each model.

\subsection{PAD performance}

\begin{figure*}[!htb]
    \begin{center}
        \subfloat[composite \label{subfig:det_co}]{{\includegraphics[width=0.33\linewidth]{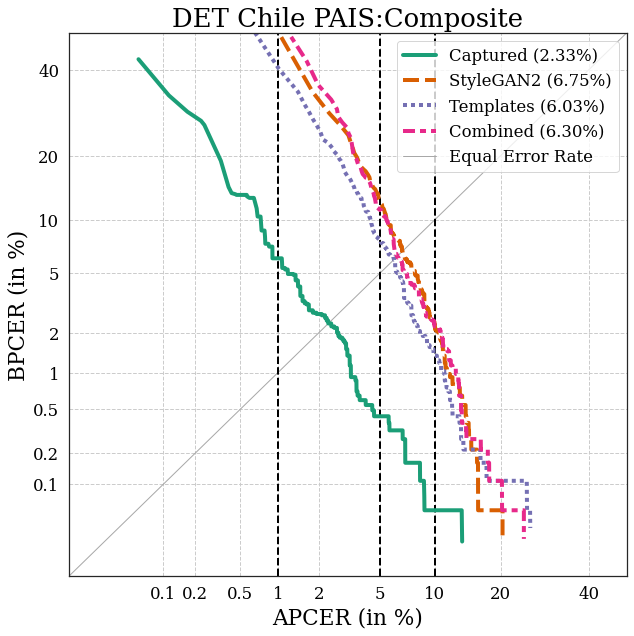}}}
        \subfloat[print \label{subfig:det_pr}]{{\includegraphics[width=0.33\linewidth]{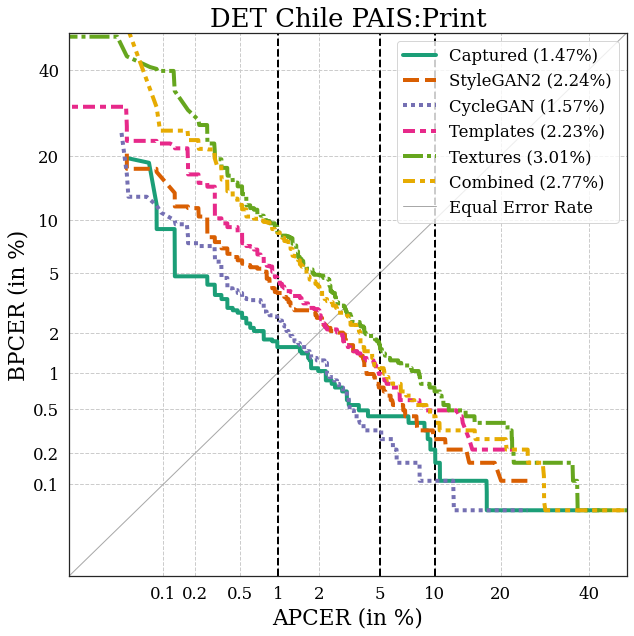}}}
        \subfloat[screen \label{subfig:det_sc}]{{\includegraphics[width=0.33\linewidth]{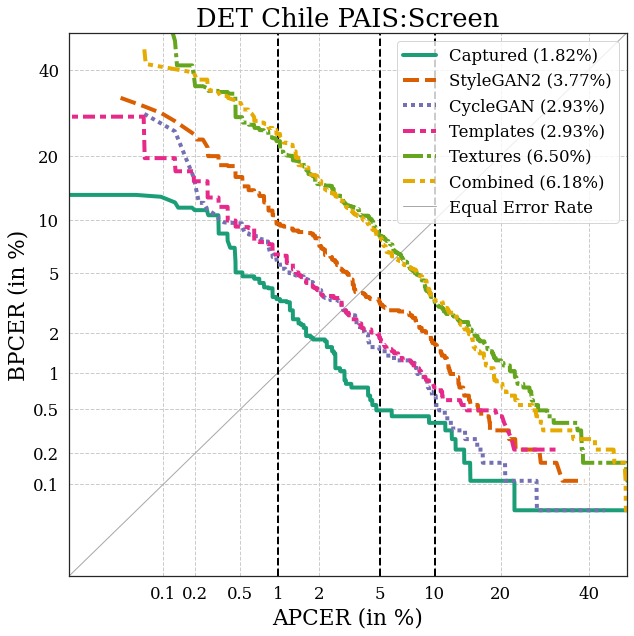}}}
    \end{center}
    \caption{Detection Error Trade-off curves for all the trained fraud-detection networks. Each plot corresponds to a different PAIS. The EER is shown in parenthesis for each scenario.}%
    \label{fig:det}%
\end{figure*}

In this work, we trained 10 networks in total, as described in Section~\ref{subsec:exp2} and Section~\ref{subsec:exp3}. The DET curves for all the networks are presented in Figure~\ref{fig:det}. Additionally, the ISO/IEC 30107-3 error rates of the ten networks are presented in Table~\ref{tab:composite} and Table~\ref{tab:source}.

\vspace{-0.3cm}

\begin{table}[!htb]
    \caption{Results for the Composite-Tampering Detection Models. SGAN2 represents StyleGAN2.}
    \centering
    \scriptsize
    \setlength{\tabcolsep}{4pt}
    \begin{tabular}{lrrrr}
        \hline
        & \textbf{Captured} & \textbf{SGAN2} & \textbf{Templ} & \textbf{Combined} \\
        \hline
        EER           & 2.33\% &  6.75\% &  6.03\% &  6.30\% \\
        BPCER\tu{10}  & 0.05\% &  2.39\% &  1.36\% &  2.33\% \\
        BPCER\tu{20}  & 0.43\% & 12.70\% &  7.82\% & 11.45\% \\
        BPCER\tu{100} & 6.13\% & 57.98\% & 43.43\% & 52.06\% \\                             
        \hline
    \end{tabular}
    \label{tab:composite}
\end{table}
\vspace{-0.3cm}

\begin{table}[!htb]
    \caption{Results for the Source-Tampering Detection Models.}
    \centering
    \scriptsize
    \setlength{\tabcolsep}{4pt}
    \begin{tabular}{lrrrrrr}
        \hline
        & \textbf{Captured} & \textbf{SGAN2} & \textbf{C-GAN} & \textbf{Templ} & \textbf{Text} & \textbf{Comb} \\
        \hline
        EER           & 1.82\% &  3.77\% & 2.93\% & 2.93\% &  6.50\% &  6.18\% \\
        BPCER\tu{10}  & 0.38\% &  1.68\% & 0.65\% & 0.76\% &  3.42\% &  3.42\% \\
        BPCER\tu{20}  & 0.49\% &  3.26\% & 1.57\% & 1.95\% &  8.25\% &  8.14\% \\
        BPCER\tu{100} & 3.53\% & 10.59\% & 6.24\% & 6.46\% & 23.51\% & 25.30\% \\                             
        \hline
    \end{tabular}
    \label{tab:source}
\end{table}

Training with captured images produces far superior performance for the composite model than mixing captured and synthetic images. The EER obtained was 2.33\%, while that of all the synthetic models was approximately 6\%, according to Figure~\ref{subfig:det_co} and Table~\ref{tab:composite}. Using Templates was marginally better than using StyleGAN2 or combining images from both methods. However, they are not suitable replacements for capturing more composite images.

For the source model, Table~\ref{tab:source} reports the scores for the worst PAI, which for all the 6 trained networks was screen. The DET curves for this PAI can be seen in Figure~\ref{subfig:det_sc}. For this PAI, using captured images produced better performance than the synthetic models; however, the performance difference is not as pronounced as in the composite model. In this case, Captured image reached an EER of 1.82\%, followed by CycleGAN and Templates with 2.93\%. Therefore, performance drops only by 1\% when the dataset is supplemented with synthetic images instead of doubling the amount of captured images. However, this is only true for CycleGAN and Templates since the other methods had EER values between 3.77\% and 6.5\%.

Although ISO/IEC 30107-3 standard indicates that only the worst-performing PAI has to be reported, we present in Figure~\ref{subfig:det_pr} the DET curves of the print PAIS for a richer comparison. In this case, captured images have a very similar performance to CycleGAN of approximately 1.5\%. This means that the labour-intensive process of printing, cutting with scissors and photographing the tampered ID cards with a smartphone can be replaced with images produced by CycleGAN without any drop in performance. The EER of the other methods is low as well. StyleGAN2 and Templates have an EER of around 2.25\%, while Textures and Combined were around 3\%. These results correlate with the findings of the FID analysis in which the print PAIS obtained the best similarity between captured and synthetic images.

Table~\ref{tab:composite} and Table~\ref{tab:source} also present the BPCER\tu{AP} values for three specific operating points. All those values follow the same trends observed with the EER.

A closer look at Figure~\ref{fig:all_methods} can reveal the subtle differences between all generation methods. StyleGAN2 could create the background, faces, signatures and composition tampering with no problems. However, it struggled with alphanumeric characters. For instance, despite proceeding from different random seeds, StyleGAN2 images of the four classes have the same first name: \enquote{Canora}. They also share the same middle name, surnames and RUN (National ID number); however, they present slight alterations that make them illegible. Reproducing legible alphanumeric characters is the most significant flaw of synthetic ID cards generated through StyleGAN2. On the other hand, Templates utilise handcrafted methods that place readable data, faces and signatures. However, the variability from image to the image was handpicked and could not represent the actual distribution of captured ID cards.

The Textures method produced excellent reproductions of the print/scan textures and pixel noise and aliasing, as seen in the four images in the lower right corner. CycleGAN generated a subtle texture for the print PAI but a coarse square texture for the screen PAI. Lastly, StyleGAN2 produced subtle textures for both print and screen. Those are closer to bona fide images than their respective attack species. 


\section{Conclusions}
\label{sec:conclusions}

This work presented four different methodologies capable of generating synthetic ID cards, and evaluated the performance of each as a possible supplement for captured images. For this purpose, we trained two MobileNetV2 networks using different combinations of captured and synthetic images. Our results indicate that the composite PAIS was the hardest to replicate. This attack scenario obtained the greatest EER and BPCER\tu{AP} compared to other PAIS. The screen PAIS produced better PAD scores than composite. For CycleGAN and Templates, there is only 1\% of performance reduction when supplementing the dataset with synthetic images instead of capturing more images. On the other hand, the print PAIS was the easiest to replicate according to the FID and PAD scores. In this case, CycleGAN achieved the same EER as using captured images only. This means supplementing with print images produced by CycleGAN would be a sound alternative to capturing more print images.

The FID analysis demonstrated that for this application, CycleGAN and Textures produced images more similar to the captured ones than those produced by StyleGAN2. These results indicate that reproducing ID card images from scratch is very demanding and challenging, and it is more feasible to use domain adaptation from bona fide images. The images from Templates achieved the worst FID scores; however, they were still valuable for training fraud detection networks, achieving reasonable PAD scores. 

Future work includes training CycleGAN using higher resolution images, improving the FID of Templates, and standardising the transferable textures for approved printers and scanners usable in Europe and USA. 

\section*{Acknowledgements}
The authors are grateful to TOC Biometrics, R\&D Center SR-226 and European Union’s Horizon 2020 research and innovation program under grant agreement 883356, and the German Federal Ministry of Education and Research and the Hessian Ministry of Higher Education, Research, Science and the Arts within their joint support of the National Research Center for Applied Cybersecurity ATHENE.

\section*{Disclaimer}
This work and the methods proposed are only for research purposes. Any implementation or commercial use modification must be analysed separately for each case to the email: juan.tapia-farias@h-da.de.

\bibliographystyle{IEEEtran}
\vspace{-0.3cm}


\bibliography{bibliography.bib}

\begin{thebibliography}{10}
\providecommand{\url}[1]{#1}
\csname url@samestyle\endcsname
\providecommand{\newblock}{\relax}
\providecommand{\bibinfo}[2]{#2}
\providecommand{\BIBentrySTDinterwordspacing}{\spaceskip=0pt\relax}
\providecommand{\BIBentryALTinterwordstretchfactor}{4}
\providecommand{\BIBentryALTinterwordspacing}{\spaceskip=\fontdimen2\font plus
\BIBentryALTinterwordstretchfactor\fontdimen3\font minus
  \fontdimen4\font\relax}
\providecommand{\BIBforeignlanguage}[2]{{%
\expandafter\ifx\csname l@#1\endcsname\relax
\typeout{** WARNING: IEEEtran.bst: No hyphenation pattern has been}%
\typeout{** loaded for the language `#1'. Using the pattern for}%
\typeout{** the default language instead.}%
\else
\language=\csname l@#1\endcsname
\fi
#2}}
\providecommand{\BIBdecl}{\relax}
\BIBdecl

\bibitem{shi2018docface}
Y.~Shi and A.~K. Jain, ``Docface: Matching id document photos to selfies,'' in
  \emph{2018 IEEE 9th International Conference on Biometrics Theory,
  Applications and Systems (BTAS)}.\hskip 1em plus 0.5em minus 0.4em\relax
  IEEE, 2018, pp. 1--8.

\bibitem{shi2019docface+}
------, ``Docface+: Id document to selfie matching,'' \emph{IEEE Transactions
  on Biometrics, Behavior, and Identity Science}, vol.~1, no.~1, pp. 56--67,
  2019.

\bibitem{gonzalez2021hybrid}
S.~González, A.~Valenzuela, and J.~Tapia, ``Hybrid two-stage architecture for
  tampering detection of chipless id cards,'' \emph{IEEE Transactions on
  Biometrics, Behavior, and Identity Science}, vol.~3, no.~1, pp. 89--100,
  2021.

\bibitem{DBLP:journals/iet-bmt/RajaR022}
R.~Mudgalgundurao, P.~Schuch, K.~B. Raja, R.~Raghavendra, and N.~Damer,
  ``Pixel-wise supervision for presentation attack detection on id cards,''
  \emph{{IET} Biometrics}, 2022.

\bibitem{Zheng2019ASO}
L.~Zheng, Y.~Zhang, and V.~L.~L. Thing, ``A survey on image tampering and its
  detection in real-world photos,'' \emph{J. Vis. Commun. Image Represent.},
  vol.~58, pp. 380--399, 2019.

\bibitem{viazovetskyi2020stylegan2}
Y.~Viazovetskyi, V.~Ivashkin, and E.~Kashin, ``Stylegan2 distillation for
  feed-forward image manipulation,'' in \emph{European Conference on Computer
  Vision}.\hskip 1em plus 0.5em minus 0.4em\relax Springer, 2020, pp. 170--186.

\bibitem{sandler2018mobilenetv2}
M.~Sandler, A.~Howard, M.~Zhu, A.~Zhmoginov, and L.-C. Chen, ``Mobilenetv2:
  Inverted residuals and linear bottlenecks,'' in \emph{Proceedings of the IEEE
  conference on computer vision and pattern recognition}, 2018, pp. 4510--4520.

\bibitem{IAN}
I.~Goodfellow, J.~Pouget-Abadie, M.~Mirza, B.~Xu, D.~Warde-Farley, S.~Ozair,
  A.~Courville, and Y.~Bengio, ``Generative adversarial nets,'' in
  \emph{Advances in Neural Information Processing Systems}, Z.~Ghahramani,
  M.~Welling, C.~Cortes, N.~Lawrence, and K.~Weinberger, Eds., vol.~27.\hskip
  1em plus 0.5em minus 0.4em\relax Curran Associates, Inc., 2014.

\bibitem{karras2020training}
T.~Karras, M.~Aittala, J.~Hellsten, S.~Laine, J.~Lehtinen, and T.~Aila,
  ``Training generative adversarial networks with limited data,''
  \emph{Advances in Neural Information Processing Systems}, vol.~33, pp.
  12\,104--12\,114, 2020.

\bibitem{chu2017cyclegan}
C.~Chu, A.~Zhmoginov, and M.~Sandler, ``Cyclegan, a master of steganography,''
  \emph{arXiv preprint arXiv:1712.02950}, 2017.

\bibitem{albiero}
V.~Albiero, N.~Srinivas, E.~Villalobos, J.~Perez-Facuse, R.~Rosenthal, D.~Mery,
  K.~Ricanek, and K.~W. Bowyer, ``Identity document to selfie face matching
  across adolescence,'' in \emph{2020 IEEE International Joint Conference on
  Biometrics (IJCB)}, 2020, pp. 1--9.

\bibitem{stokkenes2018biometric}
M.~Stokkenes, R.~Ramachandra, and C.~Busch, ``{Biometric Transaction
  Authentication using Smartphones},'' in \emph{2018 International Conference
  of the Biometrics Special Interest Group (BIOSIG)}.\hskip 1em plus 0.5em
  minus 0.4em\relax IEEE, 9 2018, pp. 1--5.

\bibitem{perera2019face}
P.~Perera and V.~M. Patel, ``{Face-Based Multiple User Active Authentication on
  Mobile Devices},'' \emph{IEEE Transactions on Information Forensics and
  Security}, vol.~14, no.~5, pp. 1240--1250, 5 2019.

\bibitem{fathy2015face}
M.~E. Fathy, V.~M. Patel, and R.~Chellappa, ``{Face-based Active Authentication
  on Mobile Devices},'' in \emph{2015 IEEE International Conference on
  Acoustics, Speech and Signal Processing (ICASSP)}.\hskip 1em plus 0.5em minus
  0.4em\relax IEEE, 4 2015, pp. 1687--1691.

\bibitem{MIDV-500}
V.~V. Arlazarov, K.~B. Bulatov, and T.~S. Chernov, ``{MIDV-500:} {A} dataset
  for identity documents analysis and recognition on mobile devices in video
  stream,'' \emph{CoRR}, vol. abs/1807.05786, 2018.

\bibitem{zhu}
X.~Zhu, H.~Liu, Z.~Lei, H.~Shi, F.~Yang, D.~Yi, G.~Qi, and S.~Z. Li,
  ``Large-scale bisample learning on {ID} versus spot face recognition,''
  \emph{Int. J. Comput. Vis.}, vol. 127, no. 6-7, pp. 684--700, 2019.

\bibitem{lara2021towards}
R.~Lara, A.~Valenzuela, D.~Schulz, J.~Tapia, and C.~Busch, ``Towards an
  efficient semantic segmentation method of id cards for verification
  systems,'' \emph{arXiv preprint arXiv:2111.12764}, 2021.

\bibitem{SalimansGZCRC16}
T.~Salimans, I.~J. Goodfellow, W.~Zaremba, V.~Cheung, A.~Radford, and X.~Chen,
  ``Improved techniques for training gans,'' \emph{CoRR}, vol. abs/1606.03498,
  2016.

\bibitem{heusel2017fid}
M.~Heusel, H.~Ramsauer, T.~Unterthiner, B.~Nessler, and S.~Hochreiter, ``Gans
  trained by a two time-scale update rule converge to a local nash
  equilibrium,'' \emph{Advances in neural information processing systems},
  vol.~30, 2017.

\bibitem{XU}
Q.~Xu, G.~Huang, Y.~Yuan, C.~Guo, Y.~Sun, F.~Wu, and K.~Q. Weinberger, ``An
  empirical study on evaluation metrics of generative adversarial networks,''
  \emph{CoRR}, vol. abs/1806.07755, 2018.

\bibitem{pouyanfar2017inception}
S.~Pouyanfar, S.-C. Chen, and M.-L. Shyu, ``An efficient deep
  residual-inception network for multimedia classification,'' in \emph{2017
  IEEE International Conference on Multimedia and Expo (ICME)}.\hskip 1em plus
  0.5em minus 0.4em\relax IEEE, 2017, pp. 373--378.

\bibitem{deng2009imagenet}
J.~Deng, W.~Dong, R.~Socher, L.-J. Li, K.~Li, and L.~Fei-Fei, ``Imagenet: A
  large-scale hierarchical image database,'' in \emph{2009 IEEE conference on
  computer vision and pattern recognition}.\hskip 1em plus 0.5em minus
  0.4em\relax Ieee, 2009, pp. 248--255.

\end{thebibliography}

\begin{IEEEbiography}[{\includegraphics[width=0.9in,height=1.25in,clip,keepaspectratio]{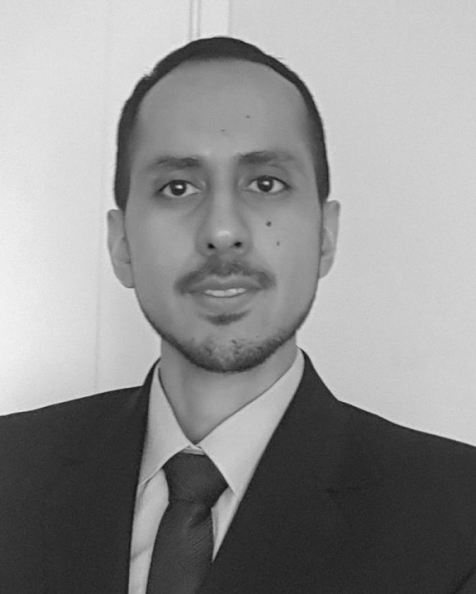}}]%
{Daniel Benalcazar}
(Member, IEEE) was born in Quito, Ecuador, in 1987. He received the B.S. degree in electronics and control engineering from Escuela Politecnica Nacional, Quito, in 2012, the M.S. degree in electrical engineering from The University of Queensland, Australia, in 2014, with a minor in biomedical engineering, and the Ph.D. degree in electrical engineering from the Universidad de Chile, Chile, in 2020. From 2015 to 2016, he worked as a Professor at the Central University of Ecuador. Ever since, he has participated in research projects in biomedical engineering and biometrics. He is currently working as a Researcher at TOC Biometrics, Chile.
\end{IEEEbiography}
\vspace{-0.3cm}

\begin{IEEEbiography}[{\includegraphics[width=0.9in,height=1.25in,clip,keepaspectratio]{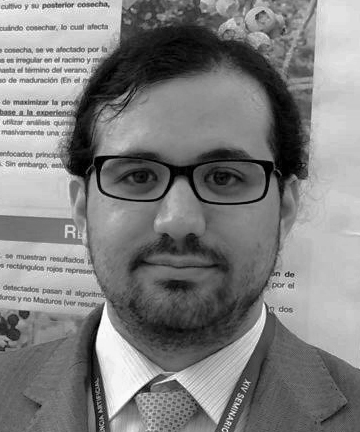}}]{Sebastian Gonzalez} received the B.Sc. degree in Computer Science from Universidad Andres Bello in 2019, and is currently pursuing a M.Sc. degree in Computer Science from Universidad de Santiago de Chile. At present, he works as a researcher at TOC Biometrics, where he has been involved in different research projects encompassing computer vision, pattern recognition, and deep learning applied to biometrics. His main interests include topics such as presentation attack detection, classification, segmentation, and applied research.
\end{IEEEbiography}
\vspace{-0.4cm}

\begin{IEEEbiography}[{\includegraphics[width=0.9in,height=1.25in,clip,keepaspectratio]{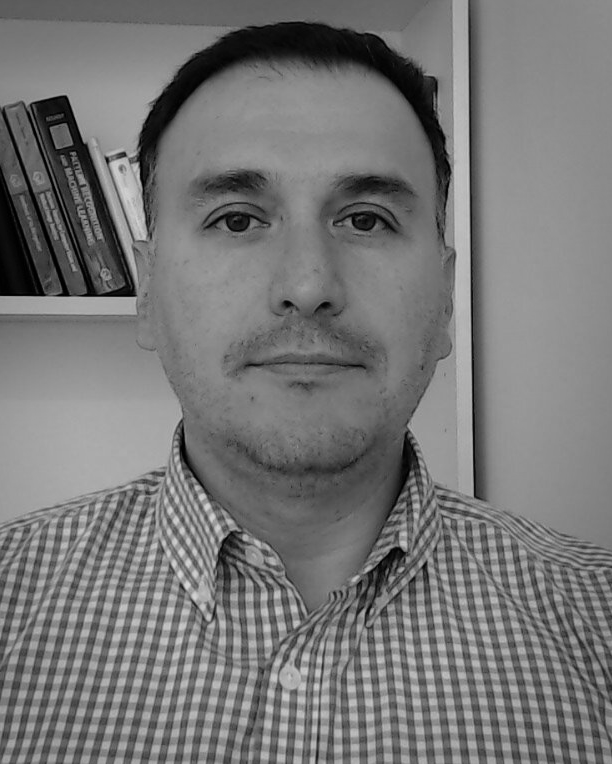}}]{Juan Tapia} received a P.E. degree in Electronics Engineering from Universidad Mayor in 2004, a M.S. in Electrical Engineering from Universidad de Chile in 2012, and a Ph.D. from the Department of Electrical Engineering, Universidad de Chile in 2016. In addition, he spent one year of internship at University of Notre Dame. In 2016, he received the award for best Ph.D. thesis. From 2016 to 2017, he was an Assistant Professor at Universidad Andres Bello. From 2018 to 2020, he was the R\&D Director for the area of Electricity and Electronics at Universidad Tecnologica de Chile. He is currently a Senior Researcher at Hochschule Darmstadt~(HDA), and R\&D Director of TOC Biometrics. His main research interests include pattern recognition and deep learning applied to iris biometrics, morphing, and feature selection. 
\end{IEEEbiography}
\vspace{-0.4cm}

\begin{IEEEbiography}[{\includegraphics[width=0.9in,height=1.25in,clip,keepaspectratio]{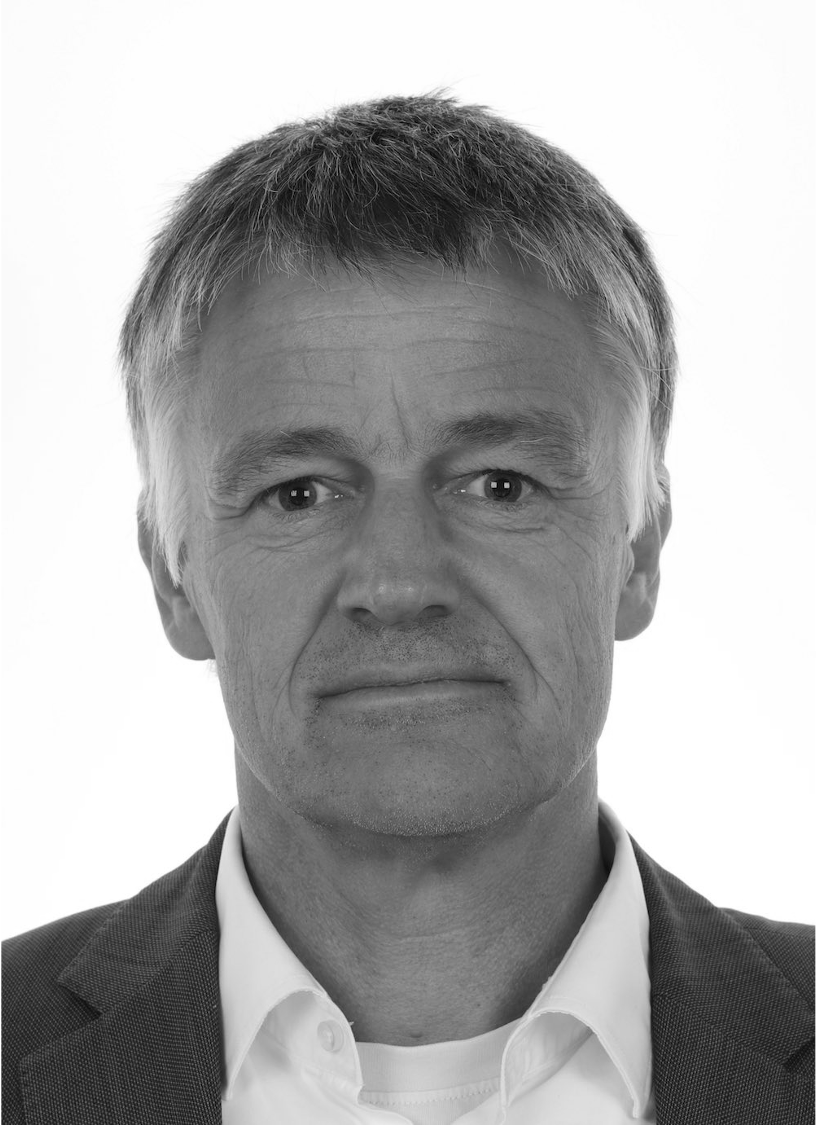}}]{Christoph Busch} is member of the Department of Information Security and Communication Technology (IIK) at the Norwegian University of Science and Technology (NTNU), Norway. He holds a joint appointment with the computer science faculty at Hochschule Darmstadt (HDA), Germany. Further he lectures the course Biometric Systems at Denmark’s DTU since 2007. On behalf of the German BSI he has been the coordinator for the project series BioIS, BioFace, BioFinger, BioKeyS Pilot-DB, KBEinweg and NFIQ2.0. In the European research program he was initiator of the Integrated Project 3D-Face, FIDELITY and iMARS. Further he was/is partner in the projects TURBINE, BEST Network, ORIGINS, INGRESS, PIDaaS, SOTAMD, RESPECT and TReSPAsS. He is also principal investigator in the German National Research Center for Applied Cybersecurity (ATHENE). Moreover Christoph Busch is co-founder and member of board of the European Association for Biometrics (www.eab.org) that was established in 2011 and assembles in the meantime more than 200 institutional members. Christoph co-authored more than 600 technical papers and has been a speaker at international conferences. He is member of the editorial board of the IET journal.
\end{IEEEbiography}

\end{document}